\documentclass[letterpaper, 10 pt, conference]{llncs}

\usepackage{graphics}    % for pdf, bitmapped graphics files
\usepackage{times}       % assumes new font selection scheme installed
\usepackage{amsmath}     % assumes amsmath package installed
\usepackage{amssymb}     % assumes amsmath package installed
\usepackage{graphicx}
\usepackage[font=small]{caption} % set table caption below Table for longer descriptions
\usepackage{wrapfig}
\usepackage{tabularx}
\usepackage{makecell}
\usepackage{multirow}
\usepackage{hhline}
\usepackage[ruled,noline,noend]{algorithm2e}
\usepackage{tikz}
\usetikzlibrary{backgrounds, fit, shapes.geometric, positioning}
\usepackage[noend]{algpseudocode}
\usepackage{url}

\def\secref#1{Sec.~\ref{#1}}
\def\figref#1{Fig.~\ref{#1}}
\def\tabref#1{Tab.~\ref{#1}}
\def\eqref#1{Eq.~(\ref{#1})}

\newcommand\etal{\emph{et al.}}
\newcommand{\comment}[1]{}

\newcolumntype{Y}{>{\centering\arraybackslash}X}% new Y, equal sized as X but centered
\title{\LARGE \bf Sensor-Based Navigation Using \\Hierarchical Reinforcement Learning}

\author{Christopher Gebauer \and Nils Dengler \and Maren Bennewitz\thanks{This work has partially been funded by the European Commission
  		under grant agreement number 964854 -- RePAIR --
                H2020-FETOPEN-2018-2020 and
                by the Deutsche Forschungsgemeinschaft (DFG, German Research Foundation) under Germany's Excellence Strategy, EXC-2070 -- 390732324 -- Phenorob. }}
\institute{Humanoid Robots Lab, University of Bonn, Germany}

\begin{document}
\maketitle
\thispagestyle{empty} 
\pagestyle{empty}

%%%%%%%%%%%%%%%%%%%%%%%%%%%%%%%%%%%%%%%%%%%%%%%%%%%%%%%%%%%%%%%%%%%%%%%%%%%%%%%%
\begin{abstract} 
  %
  % WHY is it relevant?
Robotic systems are nowadays capable of solving complex navigation tasks.
However, their capabilities are limited to the knowledge of the designer and consequently lack generalizability to initially unconsidered situations.
This makes deep reinforcement learning~(DRL) especially interesting, as these algorithms promise a self-learning system only relying on feedback from the environment.
  % WHICH PROBLEM do we address?
In this paper, we consider the problem of lidar-based robot navigation in continuous action space using DRL without providing any goal-oriented or global information.
By relying solely on local sensor data to solve navigation tasks, we design an agent that assigns its own waypoints based on intrinsic motivation.
Our agent is able to learn goal-directed navigation behavior even when facing only sparse feedback, i.e., delayed rewards when reaching the target.
To address this challenge and the complexity of the continuous action space, we deploy a hierarchical agent structure in which the exploration is distributed across multiple layers.
 % HOW is our approach special, WHAT are we actually doing, and WHAT IS NEW
Within the hierarchical structure, our agent self-assigns internal goals and learns to extract reasonable waypoints to reach the desired target position only based on local sensor data.
  %% IMPLEMENTATION, EVALUATION, WHAT FOLLOWS
In our experiments, we demonstrate the navigation capabilities of our agent in two environments and show that the hierarchical structure seriously improves the performance in terms of success rate and success weighted by path length in comparison to a flat structure.
Furthermore, we provide a real-robot experiment to illustrate that the trained agent can be easily transferred to a real-world scenario.
\end{abstract} 
%%%%%%%%%%%%%%%%%%%%%%%%%%%%%%%%%%%%%%%%%%%%%%%%%%%%%%%%%%%%%%%%%%%%%%%%%%%%%%%%
\section{Introduction}
\label{sec:intro}
%% WHY 
In the last decade, the consumer market reveals an increasing demand for mobile robots to assist a human user in a diverse set of housekeeping tasks.
While the general benefit for the consumer is out of question, most systems have a major limitation: They do not improve their behavior over time and are poor at generalization, in particular when facing situations that were not considered beforehand.
The idea of machine learning is very inspiring in that case, because it allows to search for the optimal solution and improve it over time in case of unconsidered situations or changes in expectation.
Especially, deep reinforcement learning~(DRL) is appealing as it does not require the optimal solution to learn from but only feedback from the environment.

\begin{figure}[t]
    \centering
    \includegraphics[width=0.9\linewidth]{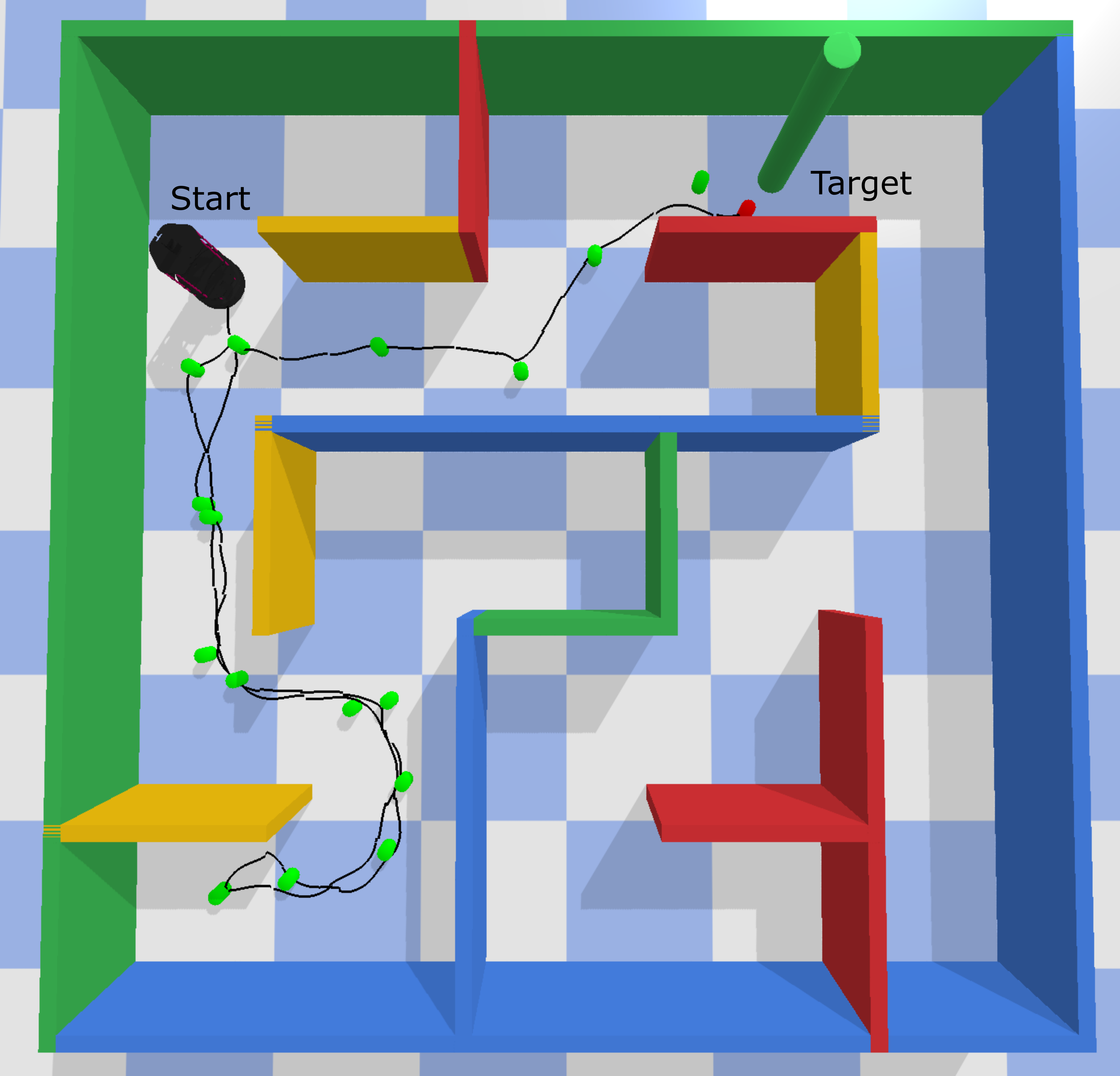}
    \caption{
        We train a hierarchical reinforcement learning agent to identify and reach its target only based on local sensor information, without providing any global or goal-oriented information. At the beginning of the training, the agent is not aware how the target looks like and where it is placed. This figure shows how the agent successfully navigates to and identifies the hidden target~(green pole) in a randomly generated maze, as indicated by the black trajectory. The small green markers represent the internal goals generated by our agent.
}
    \label{fig:env_mazes}
\end{figure}

%% WHICH PROBLEM
For robot navigation, the current state of the art uses global information about the environment in form of a map and is aware of the position of the target~\cite{sotanav}.
For reinforcement learning approaches, this information is used during runtime to provide insight to the surrounding obstacles~\cite{regier20,pfeiffer16} or the heading direction of the surrounding humans~\cite{liu20}.
In our work, we aim at designing a navigational reinforcement learning agent that is not provided with any goal-oriented or global information and that assigns its own waypoints based on intrinsic motivation, without relying on any expert knowledge.
Recently, Mirowski~\etal~\cite{mirowski17} trained a navigation agent without providing target information to the agent with very promising results.
However, the authors select the control signals from a discrete action space, which generally limits the fineness of the robot's motion capabilities.
%% HOW & WHAT 
In this paper, we present an approach to learning sensor-based navigation in \textit{continuous} action space.
To address the complexity of the continuous action space, we deploy a hierarchical agent structure~\cite{levy19,Kulkarni16nips}, which distributes the exploration across multiple layers.
The task of the agent is to navigate to a hidden target in an unknown environment, where the target is uniquely identifiable, but unknown at the beginning of the training.
We hereby consider a wheeled robot base equipped with a 2D~lidar.
To efficiently incorporate the high-dimensional information of the 2D lidar, we pretrain an autoencoder~\cite{gebauer21}, which provides a dense state representation to the hierarchical agent.

%% MAIN CONTRIBUTION & WHAT FOLLOWS FROM THAT
Our main contribution is a DRL agent capable of solving navigation tasks in continuous action space without relying on any global or target information.
As we show in the experimental evaluation, our agent learns to reach the hidden target in unknown environments solely based on local sensor data and while receiving only sparse feedback in form of delayed rewards when reaching the target.
We evaluate the performance in two different environments and show the seriously increased performance of our hierarchical structure in comparison to a flat agent regarding the success rate and success weighted by path length.
Furthermore, we transfer the trained agent into a real-world scenario and demonstrate the applicability of the learned policy on a real robot.

%%%%%%%%%%%%%%%%%%%%%%%%%%%%%%%%%%%%%%%%%%%%%%%%%%%%%%%%%%%%%%%%%%%%%%%%%%%%%%%%
\section{Related Work}
\label{sec:related}
While classical approaches for robot navigation, e.g., the dynamic window approach~\cite{fox97}, work very well in many situations~\cite{sotanav}, it is much more desirable to have algorithms that search for the optimal solution by itself~\cite{bitterlesson}.
To address this idea, Pfeiffer~\etal~\cite{pfeiffer16} used a target-driven approach that learns to steer a mobile robot collision-free to reach a relative goal pose.
Furthermore, Tai~\etal~\cite{tai17} proposed to combine reinforcement learning for local navigation with a classical global path planning algorithm to reach a target position.
Stein~\etal~\cite{stein18} on the other side, improved the global path planning by incorporating machine learning in the process of evaluating possible future subgoals.
The authors proposed to predict the expected cost for each of the possible subgoals and decrease the overall path length by choosing the most promising ones.

In contrast to the approaches above, Jaderberg~\etal~\cite{jaderberg16} introduced an agent that solves a robot navigation task end-to-end by only providing sensor information.
The agent has to explore an unknown maze to find hidden targets.
Mirowski~\etal~\cite{mirowski17} extended this approach by improving the architecture of the agent and adding a stacked long short-term memory structure.
Even though learning on raw sensor data works, Sax~\etal~\cite{sax19} showed that a well-defined state space is crucial when addressing complex navigation tasks and proposed to use the latent space of different pre-trained autoencoders.
Chaplot~\etal~\cite{chaplot20_nips} designed an object-oriented navigation agent based on semantic exploration.
This helps to understand where certain objects are more likely to be found and therefore improves the navigation capabilities.
While all these methods seem promising in general, they are limited to a discrete action space and Dhiman~\etal~\cite{dhiman18} questioned if such agents are capable of solving the navigation problem with comparable performance to classical hand-crafted approaches.
Whaid~\etal~\cite{whaid20} trained in continuous action space and compare different observations for a flat agent structure.
The authors showed that lidar information have significant impact on the performance in learning-based navigation tasks.

To increase the capabilities of an agent in complex tasks, Vezhnevets~\etal~\cite{vezhnevets17} introduced a hierarchical agent based on deep reinforcement learning, where a high-level manager sets subgoals on low frequency and a worker executes low-level actions at each time step following these subgoals.
Levy~\etal~\cite{levy19} introduced off-policy corrections based on hindsight experience replay~\cite{andrychowicz17} and reformulated the hierarchical approach to an arbitrary number of hierarchical layers.
The authors demonstrated the effectiveness in the context of robot navigation for up to three hierarchical layers.
However, in their approach the agent received the position of the goal as part of the state representation.
Li~\etal~\cite{li20} introduced a hierarchical approach to simultaneously control a robot arm and the base it is attached to.
While this again demonstrates the incredible capabilities of hierarchical structures, also here the state contains the location of the goal.
Kulkarni~\etal~\cite{Kulkarni16nips} proposed a hierarchical agent where a top-level value function learns a policy over a given set of intrinsic goals, and a lower-level function learns a policy over discrete actions to satisfy the goals.
% \mbox{Wang~\etal~\cite{wang20}} went a step further and demonstrated that a hierarchical agent is capable of solving a navigation task only relying on local sensor information.
% The authors used a model-based reinforcement learning approach to find an optimal meeting point for two mobile robots without centralized communication.

In this paper, we introduce a hierarchical reinforcement learning approach to solve the problem of sensor-based robot navigation in continuous action space in unknown environments.
% As Whaid~\etal~\cite{whaid20} showed that lidar is crucial for navigation we use this type of sensor data, even tough our formulation is not bounded to this sensor.
In contrast to the related work, we do not provide any target information within the state space and our agent identifies its own internal goals based on observations and experience from the past training.

%%%%%%%%%%%%%%%%%%%%%%%%%%%%%%%%%%%%%%%%%%%%%%%%%%%%%%%%%%%%%%%%%%%%%%%%%%%%%%%%
\section{Our Approach}
\label{sec:main}

In this section, we first introduce the navigation task to be solved.
Then, we give the preliminaries regarding reinforcement learning and the general extension to a hierarchical structure.
Next, we describe our agent and its structure in the context of the navigation task, as well as the reward function.
Finally, we discuss the countermeasures for the instabilities due to concurrent training of all layers.

\subsection{Navigation Task}
\label{sec:back}

As we explicitly avoid any global information, our agent is designed to solve general navigation problems with a designated target, 
% The target can be an arbitrary object, e.g., a lost key ring or even a hiding human, and 
which is not known to the agent at the beginning of the training.
Target information is only obtained during the training using the provided reward.
Thereby, the target needs to be uniquely recognizable by the agent from sensor data and the agent must know when the target has been reached.
%\todo{Does the data with which the VAE was trained contain the target or otherwise how is the target recognized?}
% yes.

In our application, a wheeled robot observes the external environment using a 2D~lidar sensor and measures its own state based on odometry.
We refer to these observations as $o_{\mathit{ext}}$ and $o_{\mathit{int}}$, respectively.
The robot state corresponds to a pose measurement in Cartesian space, i.e., we only consider the relative offset to the most recent measurement, thereby assuming neglectable noise.
The output of the hierarchical agent is a pair of translational and rotational velocities, which are sent to the robot base.

\begin{figure}[t]
  \centering
  \begin{tikzpicture}[
    left/.style = {shape border rotate=180}
  ]
      \node[draw, rectangle, text centered, inner sep=2mm, minimum width=8.0cm, line width=0.5mm]          (env)    at (-0.5, -0.25)       {Environment};

      \node[draw, rectangle, rounded corners, text centered, inner sep=2mm, minimum width=2.8cm]          (tfound)    at (1.8, 4.5)       {Target reached?};
      \node[draw, rectangle, rounded corners, text centered, inner sep=2mm, minimum width=2.8cm]          (gfound)    at (1.8, 3.0)       {Goal reached?};
      \node[draw, rectangle, rounded corners, text centered, inner sep=2mm, minimum width=2.8cm]          (sgfound)   at (1.8, 1.5)       {Subgoal reached?};
      \node[draw, rectangle, rounded corners, text centered, inner sep=2mm, minimum width=2.8cm]          (done)      at (1.8, 6.0)       {Done!};
      \node[draw, rectangle, rounded corners, text centered, inner sep=2mm, minimum width=2.8cm]          (init)      at (-3.0, 6.0)       {Initialize};

      \node[draw, rectangle, text centered, inner sep=2mm, line width=0.5mm]          (l3)    at (-3.0, 4.5)       {Agent L3};
      \node[draw, rectangle, text centered, inner sep=2mm, line width=0.5mm]          (l2)    at (-3.0, 3.0)       {Agent L2};
      \node[draw, rectangle, text centered, inner sep=2mm, line width=0.5mm]          (l1)    at (-3.0, 1.5)        {Agent L1};

      \draw[<-] (tfound) -- node[right, text=gray] {Yes} (gfound);
      \draw[<-] (gfound) -- node[right, text=gray] {Yes} (sgfound);
      \draw[<-] (done) -- node[right, text=gray] {Yes} (tfound);

      \draw[->] ([xshift=2mm]env.175) |- node[right] {} (l3.west);
      \draw[->] ([xshift=2mm]env.175) |- node[right] {} (l2.west);
      \draw[->] ([xshift=2mm]env.175) |- node[right, yshift=-8.8mm] {observation} (l1.west);
        
      \draw[->] (tfound.west) -- node[above, text=gray] {No} (l3.east);
      \draw[->] (gfound.west) -- node[above, text=gray] {No} (l2.east);
      \draw[->] (sgfound.west) -- node[above, text=gray] {No} (l1.east);

      \draw[->] (l3.337) -- node[right, yshift=1.5mm] {\textbf{Goal}} (gfound.170);
      \draw[->] (l2.337) -- node[right, yshift=1.5mm] {\textbf{Subgoal}} (sgfound.170);
      \draw[->] (l1.332) -- node[right] {\textbf{Command}} (env.north -| l1.332);
      \draw[<-] (sgfound.220) -- node[right] {next timestep} (env.north -| sgfound.220);

      \draw[->] (init.south) -- (l3.north);
  \end{tikzpicture}
  \caption{The information flow of our hierarchical reinforcement learning structure using one mid-level agent, i.e., $\text{I}=3$. The algorithm is initialized on the top left. The top-level agent~(L3) provides a relative goal position based on the provided observation, with the purpose to reach the hidden target. Based on the provided goal and a current observation, the mid-level agent~(L2) provides a subgoal. The low-level agent~(L1) sends commands, i.e., translational and rotational velocities, to the robot base, based on the provided subgoal and a current observation.}
  \label{fig:hrlstruct}
\end{figure}
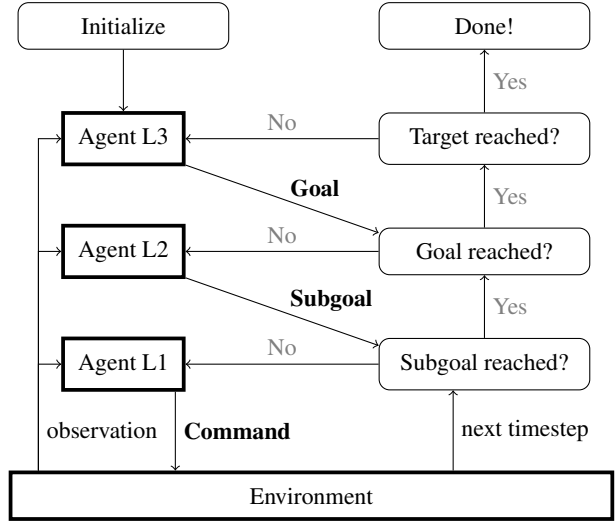 

\subsection{Preliminaries}
\label{sec:prelim}
We define our reinforcement learning setup as a partially observable Markov Decision Process~(POMDP)~\cite{pomdp} in a hierarchical structure of $\text{I}$ layers.
The hierarchical POMPD is a tuple \mbox{of~($\mathcal{S}$, $\mathcal{O}$, $\mathcal{A}$, $\mathcal{R}$, $\mathcal{T}$, $\gamma$, $\text{I}$)} components described below.
The state $s_t \in \mathcal{S}$ describes the distinct and explicit configuration of the environment in timestep $t$ and is not directly accessible for the agent.
We simplify the full hierarchical POMDP into a layer-wise \mbox{representation~($\mathcal{S}$, $\mathcal{O}_i$, $\mathcal{A}_i$, $\mathcal{R}_i$, $\mathcal{T}_i$, $\gamma$, $\Delta t_i$)}, where $i$ represents the current layer and denotes from now on layer-dependent quantities.
With this step we assume that the structure of the underlying process in each layer is identical.
Furthermore, it outlines that each layer operates on a different time scale~$\Delta t_i$, i.e., a complete episode for layer~$i$ equals one time step for layer~$i+1 \leq \text{I}$.
 
Proceeding with layer~$i$, the agent receives in timestep~$t_i$ an observation~$o_{t, i} \in \mathcal{O}_i$ and selects an \mbox{action $a_{t, i} \in \mathcal{A}_i$} according to its policy~$\pi(\theta_i): \hat{\mathcal{O}}_i \rightarrow \mathcal{A}_i$, where~$\theta_i$ parametrizes the policy.
The extended observation space~$\hat{\mathcal{O}}_i$ for~$i < \text{I}$ is computed by combining the observation space with the current action from the layer above \begin{center} $f_i : \mathcal{O}_i \times \mathcal{A}_{i+1} \rightarrow \hat{\mathcal{O}}_i$.
\end{center} For~$i = \text{I}$ it is equal to the observation space itself, i.e.,~$\hat{\mathcal{O}}_\text{I} = \mathcal{O}_\text{I}$.
After action~$a_{t, i}$ has been executed, the agent receives a reward~$r_{t, i} \in \mathcal{R}_i$ and the next observation~$o_{t+1, i}$.
The transition is selected according to the state-transition function~$\mathcal{T}_i: \mathcal{S} \times \mathcal{A}_i \times \mathcal{S}$.
The optimal policy~$\pi(\theta_i)$ maximizes the discounted reward~$R_{t_i} = \sum_{j=0}^{T_i} \gamma^j r_{t_i+j\Delta t_i}$, where~$T_i$ are the number of steps to reach the final timestep of the current episode in layer~$i$ and~$\gamma \in [0, 1]$ is the discount factor.

%\todo{which value do $\Delta t_i$ have in our application? (Exp), what is the time limit for the layers?}

\subsection{Agent Structure}
\label{sec:agentstruct}

In our application, we use a hierarchical agent with three layers as visualized in~\figref{fig:hrlstruct}.
The \textit{highest layer} addresses the navigation task with high-level abstraction, i.e., finding the hidden target within the unknown environment by proposing internal goals.
All other layers have the purpose to reach the generated internal goals.
The action for the highest layer is a relative position in Cartesian space, which we refer to as the \textit{goal}.
Since our aim is to avoid any global information, we use only the external observation as input to generate an action for Layer~3: \begin{eqnarray*} a_{t, \text{3}} = \pi(o_{\mathit{ext}, t, \text{3}}, \theta_{\text{3}} ) \quad \text{and} \quad d_{\text{3}} = p_{\text{3}}(o_{\mathit{ext}, t, \text{3}}) \end{eqnarray*} where the quantity $d_{\text{3}}$ is an indicator that signalizes whether the target has been reached, based on a predictor $p_{\text{3}}$ that is trained on past experience using the external observation.
% This either has to be provided externally, e.g., when running under supervision, or a predictor $p_{\text{I}}$ with its parameters~$\theta'_{\text{I}}$ is trained on past experience using the external observation.

The \textit{middle layer} breaks down reaching the distant goal from the highest layer to approaching closer subgoals.
In addition to the relative position in Cartesian space, this layer provides preferred velocities, i.e., to represent slowing down in narrow areas.
In the following, we will refer to the combination of both as the \textit{subgoal} provided by Layer~$2$.
As input we use the external observation but reduce the line of sight to increase the attention for nearby obstacles.
Additionally, based on the internal state and the action from the layer above, we compute a progress using the function~$f_2$, which is used as further input for the policy to generate an action.
In more detail, we determine the progress by transforming the position of the goal obtained from the layer above into the current base frame of the mobile robot, i.e., providing its current relative position.
The quantity $d_{2}$ signalizes if the relative position of the goal given from the layer above has been reached and is computed based on the progress: \begin{eqnarray*} &a_{t, 2} = \pi \left( o_{\mathit{ext}, t, 2},\mathit{prog}_2 , \theta_{2} \right) \quad \text{and} \quad d_{2} = p_{2}(\mathit{prog}_2) \\ &\text{with} \quad \mathit{prog}_2 = f_2(o_{\mathit{int}, t}, a_{t, 3}) \end{eqnarray*}

The \textit{lowest layer} directly communicates with the mobile base and sends control signals in the form of translational and rotational velocities.
It receives only the progress computed using the function~$f_1$ based on the subgoal from the layer above and the internal observation as input to generate the action: \begin{eqnarray*} &a_{t, 1} = \pi \left(\mathit{prog}_1 , \theta_{1} \right) \quad \text{and} \quad d_{1} = p_{1}(\mathit{prog}_1) \quad \\ &\text{with} \quad \mathit{prog}_1 = f_1(o_{\mathit{int}, t}, a_{t, 2}) \end{eqnarray*} The external observation is not needed as input here, because the subgoals are learned to be reachable collision-free, due to the received penalties when collisions occur, i.e., the agent automatically learns to set subgoals such that the lowest layer is capable of reaching those collision-free.
The quantity $d_{1}$ signalizes if the position of the relative position of the subgoal~$a_{t,2}$ has been reached and is computed based on the progress.

\subsection{Reward Function}
\label{sec:reward}

We divide the reward into two categories: An external reward provided from the environment $r_{\mathit{env}}$ and an intrinsic reward with two different origins, based on which layer is receiving it.
The \textit{external} reward from the environment~$r_{\mathit{env}}$ is only sparse: The robot receives a greater positive reward when it has reached the target and a smaller penalty when a collision occurs.
In all other timesteps this reward is zero.
We only provide~$r_{\mathit{env}}$ to the layers that use the external observations from the robot's sensor, i.e., Layer~3 and Layer~2.
From~$r_{\mathit{env}}$, Layer~2 learns to generate subgoals that are reachable collision-free by the lowest layer.
For Layer~2 and Layer~1 that receive the goal and subgoal, the \textit{intrinsic} reward~$r_{\mathit{prog}}$ is computed based on the progress towards the goal or subgoal.
It consists of a smaller dense reward, i.e., the distance change normalized by the initial distance to the goal or subgoal, and a greater reward when it is reached by the robot.
To have an intrinsic reward for the top layer that does not receive any goal, we grant a small reward $r_{\mathit{act}}$ correlating with the magnitude of the action, i.e., the distance to the chosen goal, to emphasize doing anything.
In our experiments, we choose the environment reward~$r_{env}$ to be a magnitude above the intrinsic motivation, as it increases the convergence speed.
The reward function is summarized as follows: \begin{equation*} 
\label{eq:rew}
r_{t,i} = \begin{cases} r_{\mathit{env}} + r_{\mathit{act}} \qquad &, \: i= 3 \\ r_{\mathit{env}} + r_{\mathit{prog}} \qquad & , \: i =2\\ r_{\mathit{prog}} \qquad &, \: i=1 \end{cases} \end{equation*} 
%\todo{which values do the rewards exactly have? (Exp)}

\subsection{Reinforcement Learning}

In this section, we first present the preprocessing of the sensor data and then describe the learning strategy for the agent.

\subsubsection{Feature Extraction} 

We use a 2D-lidar sensor as external sensor and build our preprocessing pipeline upon our previous work~\cite{gebauer21}, i.e., we convert raw lidar data into a local occupancy grid, which is encoded using a variational autoencoder~(VAE)~\cite{kingma13}.
This has two benefits in contrast to autoencoding raw lidar scans.
First, the reconstruction capabilities are significantly increased, which allows to reconstruct more accurately complex geometric shapes.
Furthermore, the grid map can be cropped without any need of replacing the new out-of-range measurements.
This reduces the information content provided to the agent when only information of nearby obstacles is required as for the middle layer of the agent~(see \secref{sec:agentstruct}) and helps decreasing the size of the latent space.
%\todo{unclear: how was the VAE trained, is target pole in trainings set?}
% yes.

\subsubsection{Learning Algorithm and Training Strategy} 

All three agents of our hierarchical learning structure are optimized with the twin delayed deep deterministic policy gradient~(TD3) algorithm~\cite{fujimoto18}.
We parametrize all policies with small dense neural networks, consisting of two layers.
%\todo{how does the network topology look like exactly?}
%check code

For training, we apply curriculum learning~\cite{bengio09} and increase the difficulty step by step, i.e., the target is placed closer and easier to reach by the robot in early stages of the training.
An episode ends, when the target has been reached, a collision occurs, or the maximal number of steps for the top layer is reached.
If one of the two other layers reaches its maximal number of steps, a new goal is provided to that layer.

Note that training the full hierarchical agent at once can lead to instabilities since even if the state-transition \mbox{probability~$\mathcal{T}_i$ for $i > 1$} is defined stationary, it will change when the policy of the layer below is adopted.
% Nachum~\etal~\cite{nachum18} did an excessive search for off-policy corrections in hierarchical reinforcement learning and showed that simply relabeling the interactions from the environment works better than any analytical solution.
Therefore, we use the hindsight actions introduced by Levy~\etal~\cite{levy19} and assume wherever the agent ended up was the intended goal position so that the reward for reaching it is experienced.
% This correction can be done either for the layer receiving the goal, to experiencing the reward for reaching it, or for the layer providing the goal, to experience an optimal lower policy.

% However, for the top layer the target cannot be changed, as it is part of the sensor data and not provided directly.
% To still have this layer experience its actions independent from the capabilities of the lower layers, we added an additional interaction, which we refer to as \textit{imaginated interaction}.
% We check what the reward would be if the desired, internal goal would have been reached.

% As we are in simulation and our reward function does not depend on the behavior of any hierarchical layer below, we can directly compute the next observation and the reward.
% We will demonstrate its effectiveness by training the top level agent only based on imagined interactions in the experimental evaluation.

In our experiments, we noticed that these additional interactions are essential at the early stage of the training, but not anymore as soon as the lower layers start to converge.
Therefore, we stop collecting these interactions when a certain reward is reached during training.

%%%%%%%%%%%%%%%%%%%%%%%%%%%%%%%%%%%%%%%%%%%%%%%%%%%%%%%%%%%%%%%%%%%%%%%%%%%%%%%%

\section{Experimental Evaluation}
\label{sec:exp}

In this section, we evaluate our approach in the context of navigation capabilities within simulation and transfer it into a real-world scenario afterwards.
At first, we specify our setup including the agent and the simulation environments.
Next, we show the performance when only training the top agent.
Finally, we compare the performance of our hierarchical agent to a flat agent and transfer the trained agent into a real-world scenario.

\begin{figure}[t]
    \centering
%    \begin{subfigure}{.59\linewidth}
        \centering
        \includegraphics[width=\linewidth]{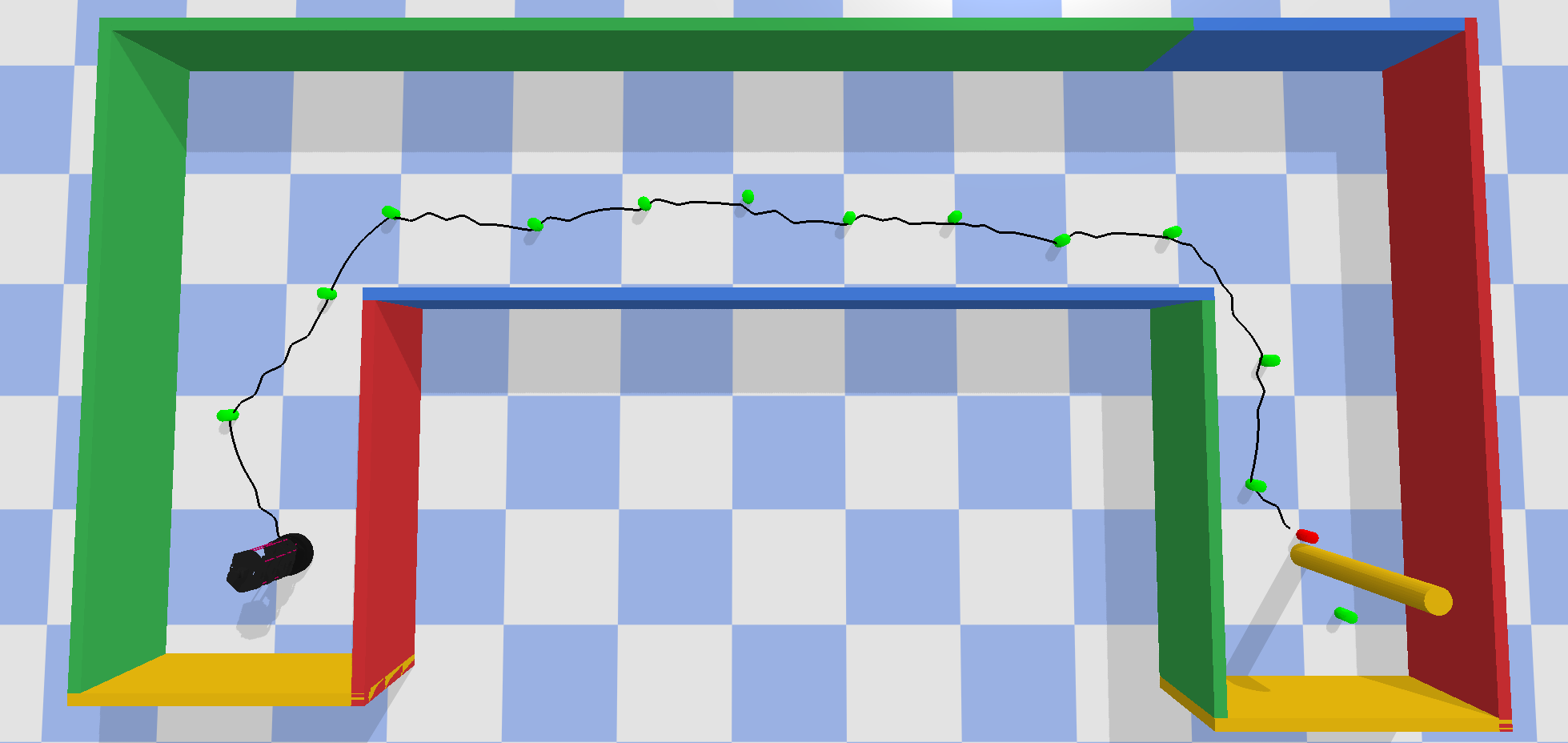}
 %   \end{subfigure}
 %   
    \caption{
      Visualization of the \textit{rooms} environment and trajectory in simulation,  the yellow pole indicates the hidden target and the green markers represent the internal goals generated by the highest layer of our agent.
}
    \label{fig:env_rooms}
\end{figure}

\subsection{Setup}
\label{sec:setup}

\subsubsection{Environment.} We test our agent in two environments with different degrees of complexity.
As external sensor we use a 2D lidar scanner with a maximal range of 10\,m.
Both environments consist of interconnected corridors or rooms with one pole representing the hidden target.
The positions of the robot and of the target are drawn randomly at the beginning of each episode.
%Furthermore, we always ensure that the robot starts away from the target.
The first environment consists of up to three interconnected rectangular rooms or corridors, see~\figref{fig:env_rooms}.
The width and length of the rooms as well as the composition are randomly drawn at the beginning of each episode.
We will refer to this environment as \textit{rooms}.
The second environment consists of a random maze, which we refer to as \textit{mazes} and which is visualized in~\figref{fig:env_mazes}.
The composition is drawn randomly, but the size is always fixed.
For the simulation we used pybullet~\cite{coumans19} and wrapped it with Gym~\cite{brockman16}.

\subsubsection{Agent.} In our implementation, the generated goal from the highest layer and the subgoal from the middle layer are represented by the relative offset in x- and y-coordinates.
For the highest layer, we use a maximal range of 1\,m and for the middle layer a range of 0.3\,m.
% We use one mid-level layer, which leads us to a total number of three layers that will be trained in parallel.
% We decided to have at least one middle layer, because Levy~\etal~\cite{levy19} showed an improved performance with an increasing number of layers.
% However, having too many layers makes the training unstable, therefore we did not increase the number of the middle layers further.
The middle layer provides a target velocity in translational direction in addition to the relative waypoint to reach in order to learn to slow down in narrow passages.
For this layer, we reduce the line of sight of the lidar sensor to 2\,m, which still includes the position of the generated goal from the top layer but already reduces the dimension of the latent space by a factor of~4.
Thus, we train an own VAE for the middle layer.
The lowest layer sends velocity commands in translational and rotational direction to the mobile base.
As time scale for the lowest layer, we use 5\,Hz.
We set the discount factor for the reward to 0.95.

\subsection{Top Layer Training}
\label{sec:topOnly}

% In this section we demonstrate the functionality of our imaginated interaction from~\secref{sec:stability} by training the top-level agent.
% We do so by only updating the top-level agent based on these interactions, while the layer below are not present.
% That is possible, because the imaginated interaction has the purpose to remove the dependency from the behavior of the layer below.

We first focus on testing different network configurations for the top-level agent. %\todo{add something from above?}
In particular, we compare the effect of concurrently continuing the training of the encoder weights with the agent and incorporating the reinforcement learning loss.
Furthermore, we investigate the influence of adding recurrent layers to the structure or the agent.
%\todo{exact architecture?} to the structure or the agent, as suggested by Mirowski~\etal~\cite{mirowski17}.

\begin{figure}[t]
    \centering 
    \includegraphics[width=0.9\linewidth]{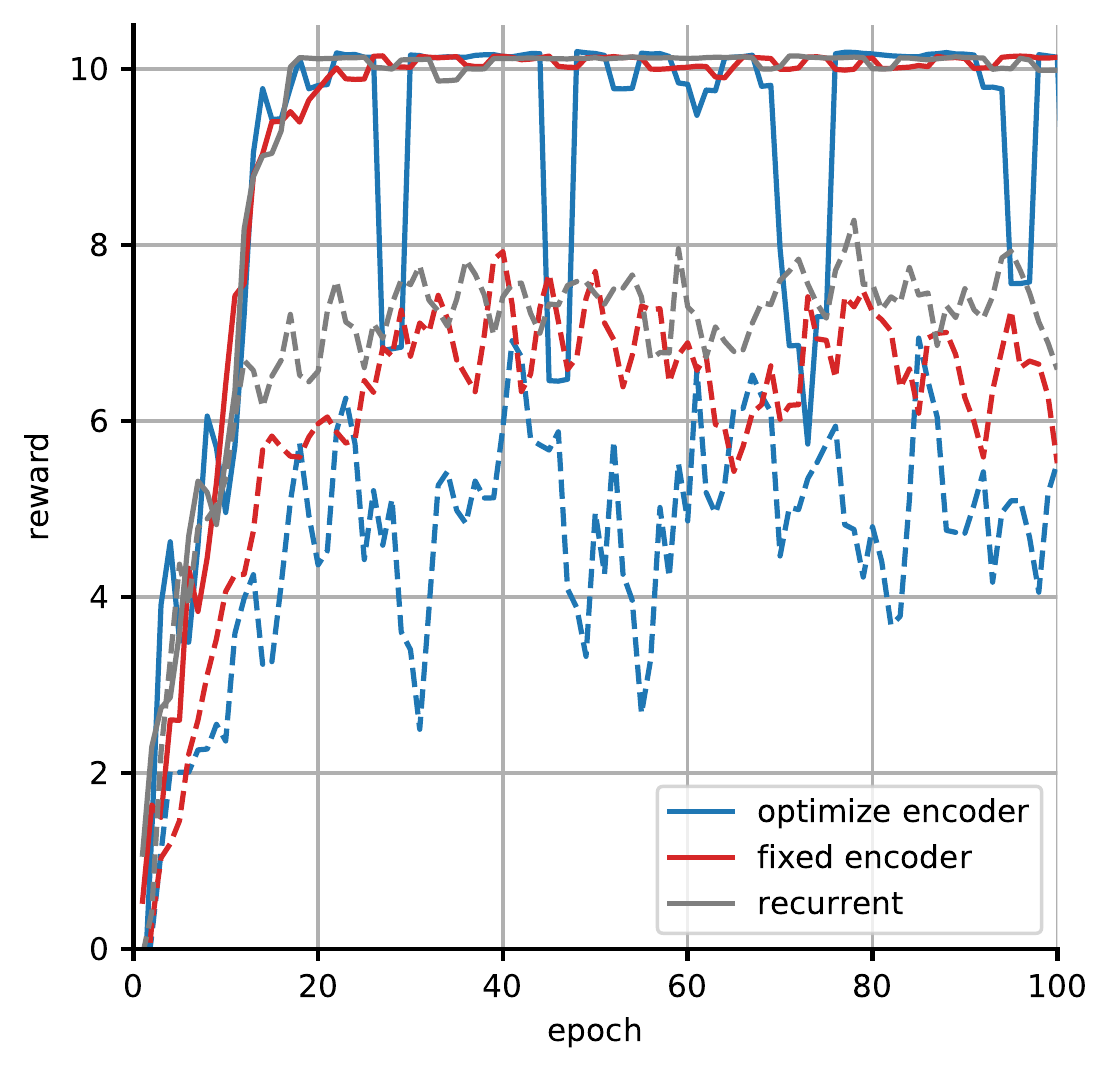}
    \caption{Comparison of different setups for the network structure of the top-level agent in the \textit{rooms}~(solid) and \textit{mazes}~(dashed) environments in terms of collected reward over the number of epochs.
Training the encoder of the lidar data concurrently~(blue) leads to instabilities, while the recurrent layer~(gray) has minimal effect. See text for a more detailed description.
All curves are averaged over three runs and smoothed with a savgol filter~\cite{savgol}.}
    \label{fig:topComp}
\end{figure}

As can be seen in~\figref{fig:topComp}, all configurations lead to similar results for the \textit{rooms} environment.
Only when training the encoder concurrently slight performance drops occur, which is originated in the fact that the agent constantly has to adapt to the changing state representation of the lidar scan.
While it is neglectable in the case of the \textit{rooms} environment, it is crucial for the \textit{mazes} environment.
Concurrently adopting to the changing state representation leads to a decreased overall performance in terms of collected reward.
However, for lidar-based agents that is no problem, because data can easily be generated from any environment upfront, i.e., an encoder can usually be pretrained and kept fixed later on.
In our previous work~\cite{gebauer21}, we even noticed great generalization capabilities, which means that retraining is usually not required.
Therefore, we will keep the encoder weights fixed. % when training the full agent in the next section.
In contrast, the recurrent agent slightly improves the performance in the \textit{mazes} environment, especially in the early stages of the training.
However, the final performance is similar and it has no effect in the \textit{rooms} environment.
Therefore, we decided for the sake of simplicity to use the dense network structure without recurrent layers for the agent from now on.

\subsection{Comparative Evaluation}

Next, we evaluate the performance of our full hierarchical structure in the two different environments.
For comparison, we train an agent % similar to Mirowski~\etal~\cite{mirowski17} but in continuous action space,
with a dense network structure without any hierarchical abstraction, 
%\todo{exact architecture?} 
which we will refer to as~\textit{flat agent}.
%\todo{was curriculum learning applied?}.
%we guess so
We use the same pretrained encoder of the laser data for the flat agent that we use for the top layer of our approach.
%, we do not apply the auxiliary losses and, as described in the previous section, only rely on a dense network for the agent.
% auxiliary losses?
As input, we pass the encoded lidar scan, the current velocity, and the last chosen action.
The output consists of velocity commands in translational and rotational direction.
The reward identical to the one of the top layer as described in~\secref{sec:reward}.

As comparison, we use the success rate of the agents.  An episode counts as success if the hidden pole is reached without collisions and before the maximal number of timesteps is reached.  From~\tabref{tab:success} it is clearly visible that our hierarchical agent structure increases the success rate in both environments compared to the flat structure.  However, when looking at the success weighted by the normalized inverse path length~(SPL)~\cite{anderson18} it becomes visible that the increased success rate comes with the price of longer, i.e., less smooth trajectories, see also the resulting trajectories in~\figref{fig:env_mazes} or in our accompanying video. As can be seen, our SPL drops in comparison to the success rate relatively more as it is the case for the flat agent. However, our SPL is still seriously higher than the SPL of the flat agent. %Therefore, our trajectories are slightly longer, which is originated in the non-smoothness.
To achieve smoother trajectories, in the future we plan to include reward terms that penalize larger orientation changes resulting from the chosen actions.

With these experiments, we have demonstrated that when using our hierarchical agent for navigation providing any global or target information can be avoided.
Furthermore, in terms of success rate we achieve an improved performance in comparison to the flat agent.
However, from the success rates in~\tabref{tab:success} it can be seen that navigation in complex environments such as \textit{mazes} is still a challenging task.

\begin{table}[t]
\centering
{\footnotesize
    \begin{tabularx}{\linewidth}{|c|Y|Y|Y|Y|}
\hline
        Environment & Metric & Flat & Our & Random Walker \\ \hline \hline
        \multirow{2}{*}{Rooms} & Success & 0.664 & \textbf{0.969} & 0.016 \\ \cline{2-5}
                               & SPL~\cite{anderson18} & 0.573 & \textbf{0.782} & 0.015 \\ \hline
        \multirow{2}{*}{Mazes} & Success & 0.342 & \textbf{0.581} & 0.003 \\ \cline{2-5}
                               & SPL~\cite{anderson18} & 0.306 & \textbf{0.452} & 0.003 \\ \hline
\end{tabularx}
}
\caption{
    Success rate and success weighted by the normalized inverse path length~(SPL)~\cite{anderson18} for the \textit{rooms} and \textit{mazes} environment. As can be seen, our agent seriously increases the success rate and the SPL in comparison to the flat agent. The random walker, executing randomly drawn actions, is more than one magnitude below the flat and our agent. % The reduced smoothness of our generated trajectory becomes noticeable, when looking at the SPL. Our SPL drops in comparison to the success rate relatively more than for the flat agent. Even though, SPL takes the success rate into account the non-smoothness leads to a longer trajectory and therefore alleviates the improved success rate. Furthermore, the lower success rates for \textit{mazes} compared to \textit{rooms} show that complex environments are still challenging when avoiding any global or target information.
    All values are obtained by averaging over 1000 trials.
}
\label{tab:success}
\end{table}

\begin{figure}[t]
%   \begin{subfigure}{.395\linewidth}
        \centering
        \includegraphics[width=11cm]{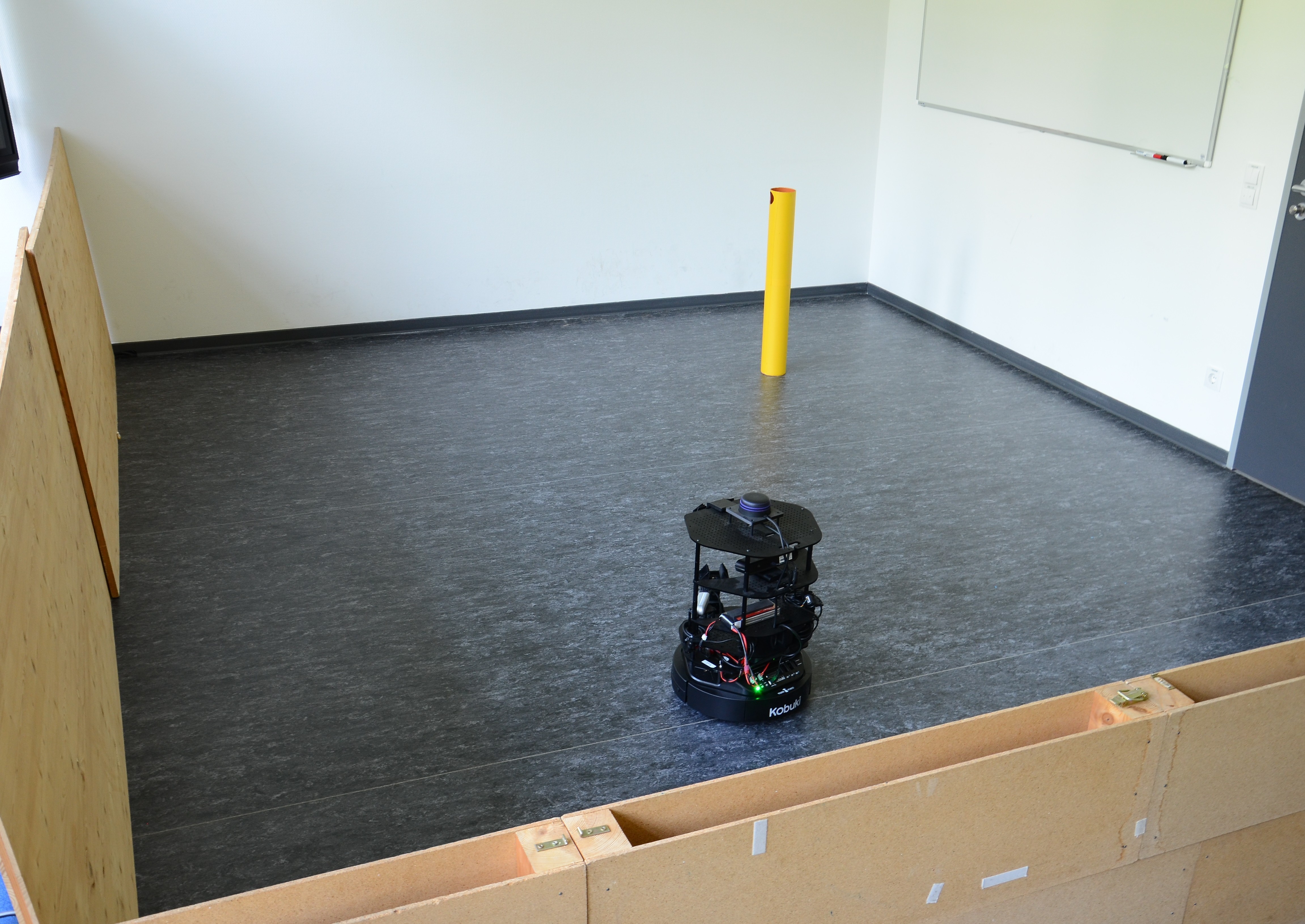}
 %   \end{subfigure}
    \caption{
        Real-world test environment in our lab where the yellow pole represents the target to reach. 
}
    \label{fig:env_real}
\end{figure}

\subsection{Transfer to Real-World}

Finally, we evaluate the transferability of our hierarchical agent to a real-world scenario after training in simulation.
Note that neither the encoder nor the agent itself have been trained on real-world data.
For evaluation we built a setup similar to the \textit{rooms} environment, see~\figref{fig:env_real}.
As mobile base we used a TurtleBot2i that was also used in simulation.
We placed the target and the robot on 15 different positions and started our hierarchical algorithm.
To unambiguously determine the end of an episode, we let the algorithm run until the robot touched the target.
In each episode, our agent successfully reached the target, i.e., we achieved a success rate of 1.0 and an SPL of 0.742.
In simulation, our agent reached in a room with the shape of the real environment in~\figref{fig:env_real}, also an success rate of 1.0 and an SPL of 0.738 averaged over 1000 trials.
Thus, the results obtained in the real-world scenario are comparable to the ones from simulation.

%%%%%%%%%%%%%%%%%%%%%%%%%%%%%%%%%%%%%%%%%%%%%%%%%%%%%%%%%%%%%%%%%%%%%%%%%%%%%%%%
\section{Conclusion}
\label{sec:conclusion}

In this paper, we presented a hierarchical structure for learning navigation tasks using deep reinforcement learning.
Our agent learns to reach a hidden target in unknown environments by self-assigning internal goals and choosing waypoints to reach the target position only based on local sensor data.
In this way, we avoid providing any global or goal-directed information and generate an intrinsically motivated agent that discovers its own targets while receiving only a delayed reward.
To exploit the full capabilities of a mobile robot, we use the continuous action space of target velocities, which also simplifies the transfer to real-world scenarios.

In experiments with a wheeled base equipped with a 2D~lidar, we showed that our agent is able to solve navigation tasks using only local sensor data and demonstrated that the hierarchical structure clearly outperforms a flat agent in terms of success rate and success weighted by path length in two different simulated environments.
Furthermore, we successfully transferred the trained agent to a real-world scenario.

%%%%%%%%%%%%%%%%%%%%%%%%%%%%%%%%%%%%%%%%%%%%%%%%%%%%%%%%%%%%%%%%%%%%%%%%%%%%%%%%

\bibliographystyle{splncs03}

\bibliography{bibliography}

\end{document}